\let\accentvec\vec 
\documentclass[runningheads,a4paper,envcountsame]{llncs}
\usepackage[report]{modstruc}
\usepackage{longtable}

\usepackage[caption=false]{subfig} 
\let\vec\accentvec
\usepackage{amssymb}
\usepackage{amsmath}
\usepackage{float}
\usepackage{url}

\floatstyle{ruled}
\newfloat{listing}{thp}{lop}
\floatname{listing}{Listing}

\begin{document}

\mainmatter  

\title{%
  Syntactic vs.\ Semantic Locality:\\
  How Good Is a Cheap Approximation?
}


%
%
\author{Chiara Del Vescovo\inst{1}\and Pavel Klinov\inst{2}\and Bijan Parsia\inst{1}\and\\ Uli Sattler\inst{1}\and Thomas Schneider\inst{3}\and Dmitry Tsarkov\inst{1}}

%
\authorrunning{C. Del Vescovo, P. Klinov, B. Parsia, U. Sattler, T. Schneider, D. Tsarkov}

\institute{
University of Manchester, UK \\
\url{{delvescc,bparsia,sattler,tsarkov}@cs.man.ac.uk}
\and
University of Ulm, Germany\\
\url{pavel.klinov@uni-ulm.de} \\
\and
Universit\"{a}t Bremen, Germany \\
\url{tschneider@informatik.uni-bremen.de} \\
}

\maketitle

\begin{abstract}
  Extracting a subset of a given OWL ontology that captures all the ontology's knowledge about a specified set of terms is a well-understood task.
  This task can be based, for instance, on locality-based modules (LBMs).
  These come in two flavours, syntactic and semantic, and a syntactic LBM is known to contain the corresponding semantic LBM.
  For syntactic LBMs, polynomial extraction algorithms are known, implemented in the OWL API, and being used.
  In contrast, extracting semantic LBMs involves reasoning, which is intractable for OWL 2 DL, and these algorithms had not been implemented yet
  for expressive ontology languages.
  
  We present the first implementation of semantic LBMs and report on experiments that compare them with syntactic LBMs
  extracted from real-life ontologies. 
  Our study reveals whether semantic LBMs are worth the additional extraction effort, compared with syntactic LBMs.
\end{abstract}

\section{Introduction}  

Extracting a subset of a given OWL ontology that captures all
the ontology's knowledge about a specified set of concept and role
names is an interesting task for various applications, and it is by now
well-understood 
\cite{CHKS08,KLWW08a,KWZ10}. In general, we
consider a setting where, for a given \emph{signature}, we want to
determine a (small) subset of a given ontology such that any axiom
over the signature entailed by the ontology is also entailed by the
subset. For expressive logics, this task can be implemented by making use of the
notion of \emph{locality}, and results in what is known as
locality-based modules (LBMs) \cite{CHKS08}.
Locality comes in many different flavours, in particular there are
notions of syntactic and semantic locality. A syntactic LBM is known
to contain the corresponding semantic LBM, but might also contain
extra axioms which are, because they are not in the semantic LBM,
superfluous for entailments over the given signature.  Algorithms for
the extraction of syntactic LBMs are known that run in time that is
polynomial in the size of the ontology (thus much cheaper than
reasoning), implemented in the OWL API, and being used.  In contrast,
despite the fact that algorithms for 
extracting semantic LBMs are known, until now and to the best of our knowledge, they had not yet been
implemented. Moreover, these involve entailment checking, and are thus intractable for expressive profiles of 
OWL 2. 
  
We present the first implementation of semantic LBMs and report on experiments that compare them with syntactic LBMs
extracted from real-life ontologies. The contributions of this paper
are as follows: we show with statistical significance that, for almost all members of a large corpus of existing ontologies,
there is no difference between any syntactic LBM and its corresponding semantic LBM.
In the few cases where differences occur, these differences are modest and not worth the increased
computation time needed to compute semantic LBMs. In addition,
we isolate two types of axioms that lead to differences,
where one is a simple tautology that can, in principle,
be detected by a straightforward addition to the syntactic locality checker.
Furthermore, our results show that the extraction of semantic LBMs,
which is in principle hard, seems feasible in practice.
The lesson we learn from these results is that ``Cheap is Great''!

\section{Preliminaries}
\label{sec:prelims}
We assume the reader to be familiar with OWL and the underlying description logic \DL{SROIQ}~\cite{DLHB03,HoKS06},
and will define the central notions around locality-based modularity~\cite{CHKS08}.

Let \Terms{C} be a set of concept names, and \Terms{R} a set of role names.
A \emph{signature} $\Sigma$ is a set of \emph{terms}, i.e., a set $\Sigma \subseteq \Terms{C} \cup \Terms{R}$ of concept and role names.
We can think of a signature as specifying a topic of interest.
Axioms that only use terms from $\Sigma$ can be thought of as ``on-topic'', and all other axioms as ``off-topic''.
For instance, if $\Sigma = \{\Term{Animal}, \Term{Duck}, \Term{Grass}, \Term{eats}\}$,
then $\Term{Duck} \sqsubseteq \exists \Term{eats}.\Term{Grass}$ is on-topic,
while $\Term{Duck} \sqsubseteq \Term{Bird}$ is off-topic.

Any concept, role, or axiom that uses only terms from $\Sigma$ is called a \emph{$\Sigma$-concept},
\emph{$\Sigma$-role}, or \emph{$\Sigma$-axiom}.
Given any such object $X$, 
we call the set of terms in $X$ the \emph{signature of $X$}
and denote it with $\Sig{X}$.

Given an interpretation \Int{I}, we denote its restriction to the terms in a signature $\Sigma$
with $\Int{I}|_\Sigma$.
Two interpretations \Int{I} and \Int{J} are said to \emph{coincide on a signature $\Sigma$},
in symbols $\Int{I}|_\Sigma = \Int{J}|_\Sigma$, if $\Delta^{\Int{I}} = \Delta^{\Int{J}}$
and $X^{\Int{I}} = X^{\Int{J}}$ for all $X \in \Sigma$.

There are a number of variants of the notion of conservative extensions, which
capture the desired preservation of knowledge to different degrees.
We focus on the deductive variant.

\begin{defi}
  \label{def:CE+modules}
  Let $\module \subseteq \ontol$ be \DL{SROIQ}-ontologies
  and $\Sigma$ a signature.
  \begin{Enum}
    \item
      $\ontol$ is a \emph{deductive $\Sigma$-conservative extension \textup{(}$\Sigma$-dCE\textup{)}} of $\module$
      if, for all \DL{SROIQ}-axioms $\alpha$ with $\Sig{\alpha} \subseteq \Sigma$, it holds that
      $\module \models \alpha$ if and only if $\ontol \models \alpha$.
    \item
      \label{it:CE-based_modules}
      $\module$ is a \emph{dCE-based module for $\Sigma$} of $\ontol$ if
      $\ontol$ is a $\Sigma$-dCE of $\module$.
  \end{Enum}
\end{defi}


Unfortunately, deciding in general if a set of axioms is a module in this sense is hard or even impossible for expressive DLs~\cite{GhLW06,LuWW07}, and
finding a minimal one is even more so. However, ``good sized'' modules that are efficiently computable have been introduced~\cite{CHKS08}.
They are based on the \emph{locality} of single axioms, which means that, given $\Sigma$,
the axiom can always be satisfied independently of the interpretation of the $\Sigma$-terms,
but in a restricted way: by interpreting all non-$\Sigma$ terms either as the empty set ($\emptyset$-locality)
or as the full domain\footnote{%
  Or, in the case of roles, the set of all pairs of domain elements.
} ($\Delta$-locality).

\begin{defi}
  A \DL{SROIQ}-axiom $\alpha$ is called 
  \emph{$\emptyset$-local ($\Delta$-local) w.r.t.\ signature $\Sigma$}
  if, for each interpretation $\Int{I}$, there exists an interpretation
  $\Int{J}$ such that $\Int{I}|_\Sigma = \Int{J}|_\Sigma$, $\Int{J} \models \alpha$,
  and for each $X \in \Sig{\alpha} \setminus \Sigma$, $X^{\Int{J}} = \emptyset$
  (for each $C \in \Sig{\alpha} \setminus \Sigma$, $C^{\Int{J}} = \Delta$
  and for each $R \in \Sig{\alpha} \setminus \Sigma$, $R^{\Int{J}} = \Delta \times \Delta$).
\end{defi}

It has been shown in~\cite{CHKS08} that $\TBox{M} \subseteq\, \TBox{O}$
and all axioms in $\TBox{O}\setminus\TBox{M}$ being $\emptyset$-local
(or all axioms being $\Delta$-local) w.r.t.\ $\Sigma \cup \Sig{\TBox{M}}$
is sufficient for \TBox{O} to be a $\Sigma$-dCE of \TBox{M}.
The converse does not hold: e.g., the axiom $A \equiv B$ is neither
$\emptyset$- nor $\Delta$-local w.r.t.\ $\{A\}$, but the ontology
$\{A \equiv B\}$ is an $\{A\}$-dCE of the empty ontology.
\sloppy

Furthermore, locality can be tested using available DL-reasoners~\cite{CHKS08},
which makes this problem considerably easier than testing conservativity.
However, reasoning in expressive DLs is still complex, e.g.\
\CC{N2Exp\-Time}-complete for \DL{SROIQ}~\cite{Kaz08}.
In order to achieve \emph{tractable} module extraction, a syntactic approximation
of locality has been introduced in~\cite{CHKS08}.
The following definition captures only the case of \DL{SHQ}-TBoxes and can straightforwardly be extended to \DL{SROIQ} ontologies.

\fussy
\begin{defi}
  An axiom $\alpha$ is called
  \emph{syntactically $\bot$-local \textup{(}$\top$-local\textup{)} w.r.t.\ signature $\Sigma$}
  if it is of the form $C^\bot \sqsubseteq C$, $C \sqsubseteq C^\top$, 
  $C^\bot \equiv C^\bot$, $C^\top \equiv C^\top$,
  $R^\bot \sqsubseteq R$ ($R \sqsubseteq R^\top$), or
  $\Trans(R^\bot)$ ($\Trans(R^\top)$),
  where $C$ is an arbitrary concept, $R$ is an arbitrary role name,
  $R^\bot \notin \Sigma$ ($R^\top \notin \Sigma$), and
  $C^\bot$ and $C^\top$ are from $\Bot(\Sigma)$ and $\Top(\Sigma)$
  as defined in
  Part (a) (resp.\ (b)) of the table below.
\end{defi}
  \begin{small}
    \begin{center}
    \newcommand{\nplus}{\bar{n}}
    \newcommand{\dnot}{\neg}
    \newcommand{\dand}{\sqcap}
    \newcommand{\dex}[2]{\exists #1.#2}
    \newcommand{\atleastq}[3]{\mathord\geqslant #1\,#2.#3}
    \newcommand\rrule{\rule[-1.3ex]{0pt}{3.5ex}}
    \begin{tabular}[t]{@{}>{\rrule$}l<{$}@{\;}>{$}l<{$}@{}}
      \multicolumn{2}{@{}l@{}}{%
        \rule[-5pt]{0pt}{10pt}
        (a)~$\bot$-\textsl{Locality} \hspace{7mm}
        Let $A^\bot,R^\bot \notin \Sigma,~ C^{\bot}\in\Bot(\Sigma)$,~ $C_{(i)}^{\top}\in\Top(\Sigma),~ \nplus\in\mathbb{N}\setminus\{0\}$%
      }                                                                                                                             \\
      \hline\rule{0pt}{11pt}%
      \Bot(\Sigma) & ::= A^{\bot} \mid \bot \mid  \dnot C^{\top}  \mid  C \dand C^{\bot}  \mid  C^{\bot} \dand C
                        \mid \dex{R}C^{\bot}\mid  \atleastq{\nplus}{R}{C^{\bot}} \mid \dex{R^{\bot}}C 
                        \mid \atleastq{\nplus}{R^{\bot}}C                                                                       \\
      \Top(\Sigma) & ::= \top \mid \dnot C^{\bot} \mid C_1^{\top}\dand C_2^{\top} \mid \atleastq{0}{R}{C} \\
      \hline
    \end{tabular}
    \par\bigskip
    \begin{tabular}[t]{@{}>{\rrule$}l<{$}@{\;}>{$}l<{$}@{}}
      \multicolumn{2}{@{}l@{}}{%
        \rule[-5pt]{0pt}{10pt}
        (b)~$\top$-\textsl{Locality} \hspace{7mm}
        Let $A^\top,R^\top \notin \Sigma,~ C^{\bot}\in\Bot(\Sigma)$,~ $C_{(i)}^{\top}\in\Top(\Sigma),~ \nplus\in\mathbb{N}\setminus\{0\}$%
      }                                                                                                                            \\
      \hline\rule{0pt}{11pt}%
      \Bot(\Sigma) & ::= \bot \mid \dnot C^{\top} \mid C \dand C^{\bot} \mid C^{\bot} \dand C
                      \mid \dex{R}C^{\bot}\mid \atleastq{\nplus}{R}{C^{\bot}}                                                \\
      \Top(\Sigma) & ::= A^{\top} \mid \top \mid \dnot C^{\bot} \mid C_1^{\top}\dand C_2^{\top}
                      \mid \dex{R^{\top}}C^{\top} \mid \atleastq{\nplus}{R^{\top}}{C^{\top}} \mid \atleastq{0}{R}{C}    \\
      \hline
    \end{tabular}

  \end{center}
  \par
\end{small}

It has been shown in~\cite{CHKS08} that $\bot$-locality ($\top$-locality) of an axiom $\alpha$
w.r.t.\ $\Sigma$ implies $\emptyset$-locality ($\Delta$-locality) of $\alpha$ w.r.t.\ $\Sigma$.
Therefore, all axioms in $\TBox{O}\setminus\TBox{M}$ being $\bot$-local
(or all axioms being $\top$-local) w.r.t.\ $\Sigma \cup \Sig{\TBox{M}}$
is sufficient for \TBox{O} to be a $\Sigma$-dCE of \TBox{M}.
The converse does not hold; examples can be found in~\cite{CHKS08}.

For each of the four locality notions, modules of \TBox{O} are obtained by starting with an empty set of axioms and subsequently
adding axioms from \TBox{O} that are $\Sigma$-non-local.
In order for this procedure to be correct, the signature against which locality
is checked has to be extended with the terms in the axioms that are added in each step, so that the resulting module $\module$ consists of all
the non-local axioms with respect to $\Sigma\cup\Sig{\module}$.
Definition~\ref{def:syn_loc_modules}\,\ref{it:x-module} introduces locality-based modules,
which are always dCE-based modules~\cite{CHKS08}, although not necessarily minimal ones.
Modules based on syntactic (semantic) locality can be made smaller by iteratively nesting $\top$- and $\bot$-extraction ($\semtop$- and
$\sembot$-extraction),
and the result is still a dCE-based module~\cite{CHKS08,SaSZ09}.
These so-called $\topbot^\ast$-modules ($\semtop\sembot^\ast$-modules) are introduced in Definition~\ref{def:syn_loc_modules}\,\ref{it:yz-star-module}.

\begin{defi}
  \label{def:syn_loc_modules}
  Let $x \in \{\emptyset,\Delta,\bot,\top\}$, $yz \in \{\topbot,\semtop\sembot\}$, \TBox{O} an ontology and $\Sigma$ a signature.
  \begin{Enum}
    \item
      \label{it:x-module}
      An ontology \TBox{M} is the \emph{$x$-module of \TBox{O} w.r.t.\ $\Sigma$}
      if it is the output of Algorithm~\ref{alg:x-module}.
      We write $\TBox{M} = \xmod(\Sigma,\TBox{O})$.
    \item
      \label{it:yz-module}
      An ontology \TBox{M} is the \emph{$yz$-module of \TBox{O} w.r.t.\ $\Sigma$},
      written $\TBox{M} = \yzmod(\Sigma,\TBox{O})$,
      if $\TBox{M} = \ymod(\Sigma, \zmod(\Sigma, \TBox{O}))$.
    \item
      \label{it:yz-star-module}
      Let $(\TBox{M}_i)_{i \geqslant 0}$ be a sequence of ontologies such that
      $\TBox{M}_0 = \TBox{O}$ and $\TBox{M}_{i+1} = \yzmod(\Sigma, \TBox{M}_i)$
      for every $i \geqslant 0$. For the smallest $n \geqslant 0$ with
      $\TBox{M}_n = \TBox{M}_{n+1}$, we call $\TBox{M}_n$ the
      \emph{$yz^\ast$-module of \TBox{O} w.r.t.\ $\Sigma$},
      written $\TBox{M} = \yzsmod(\Sigma,\TBox{O})$.
  \end{Enum}
\end{defi}
\begin{algorithm}
  \begin{algorithmic}
    \vspace*{2pt}
    \STATE{%
      \textbf{Input:}~  Ont. \TBox{O},~ sig. $\Sigma$,~ $x \in \{\emptyset,\Delta,\bot,\top\}$\hfill
      \textbf{Output:}~ $x$-module \TBox{M} of \TBox{O} w.r.t.\ $\Sigma$%
    }
    \vspace*{4pt}
    \hrule
    \vspace*{4pt}
    \STATE{$M \leftarrow \emptyset$;~ $\TBox{O}' \leftarrow \TBox{O}$}
    \REPEAT
      \STATE{changed $\leftarrow$ \texttt{false}}
      \FORALL{$\alpha \in \TBox{O}'$}
        \vspace*{2pt}
        \IF{$\alpha$ not $x$-local w.r.t.\ $\Sigma \cup \Sig{\TBox{M}}$}
          \STATE{$\TBox{M} \leftarrow \TBox{M} \cup \{\alpha\}$;~ $\TBox{O}' \leftarrow \TBox{O}' \setminus \{\alpha\}$;~ changed $\leftarrow$ \texttt{true}}
        \ENDIF
      \ENDFOR
    \UNTIL{changed $=$ \texttt{false}}
    \STATE{\textbf{return} \TBox{M}}
  \end{algorithmic}
  \caption{Extract a locality-based module}
  \label{alg:x-module}
\end{algorithm}

As for~\ref{it:x-module},
it has been shown in~\cite{CHKS08} that the output \TBox{M}
of Algorithm~\ref{alg:x-module} does not
depend on the order in which the axioms $\alpha$ are selected.\footnote{%
  Our algorithm is a special case of the one in~\cite[Figure 4]{CHKS08}.%
}
Furthermore, the integer $n$ in
\ref{it:yz-star-module}
exists because the sequence $(\TBox{M}_i)_{i \geqslant 0}$ is
decreasing (more precisely, we have
$\TBox{M}_0 \supset \dots \supset \TBox{M}_n = \TBox{M}_{n+1} = \dots$).
Due to monotonicity properties of locality-based modules,
the dual notions of $\bottop^\ast$- and $\sembot\semtop^\ast$-modules are uninteresting
because they coincide with those of $\topbot^\ast$- and $\semtop\sembot^\ast$-modules.

Roughly speaking, a $\semtop$- or $\top$-module for $\Sigma$ gives a view
from above because it contains all subclasses of class names in $\Sigma$, while a $\sembot$- or $\bot$-module for $\Sigma$ gives a view from below since it contains all
superconcepts of concept names in $\Sigma$.

Modulo the locality check, Algorithm~\ref{alg:x-module}
runs in time cubic in $|\TBox{O}| + |\Sigma|$~\cite{CHKS08}.
Modules based on $\bot$/$\top$-locality are therefore
a feasible approximation for modules based on $\emptyset$/$\Delta$-locality.
In both cases, modules are extracted axiom by axiom but, as said above, the $\emptyset$/$\Delta$-locality check is more complex.
A module extractor is implemented in the OWL API\footnote{\url{http://owlapi.sourceforge.net}} and SSWAP\footnote{\url{http://sswap.info}}.
To summarize:
\begin{enumerate} 
\item Given an ontology $\ontol$, the semantic module $\semmodule_\Sigma$ for a signature $\Sigma$ is contained in the corresponding syntactic module
$\synmodule_\Sigma$ for the same seed signature.\footnote{Recall that $\bot$-syntactic modules approximate $\emptyset$-semantic modules, while
$\top$-syntactic modules approximate $\Delta$-semantic modules.} This means that in principle more unnecessary axioms for preserving entailments over
$\Sigma$ can end up in syntactic modules rather than in semantic modules.

\item The extraction of a syntactic module can be done in polynomial time w.r.t.\ the size of the ontology $\ontol$. In contrast, the extraction of a semantic
module is as hard as reasoning. 
\end{enumerate}

\section{Experimental design}
\label{sec:expDesign}

The main aim of this paper is to investigate how well syntactic locality approximates semantic locality. In particular, we want to see how (un)likely it is that syntactic locality-based modules are larger than semantic locality-based ones and how large these differences are. We also want to understand empirically how much more costly semantic locality is in terms of performance.

\paragraph*{Selection of the Corpus.}

For our experiments, we have built a corpus containing: ($1$) from the TONES repository,\footnote{\url{http://owl.cs.manchester.ac.uk/repository/}} those ontologies that have already been studied in a previous work on modularity~\cite{DPSS10womo}: \Koala, \Mereology, \University, \People, \miniTambis, \OWLS, \Tambis, \Galen; ($2$) all ontologies from the NCBO BioPortal ontology repository.\footnote{\url{http://bioportal.bioontology.org}}  

We then filter out all those the ontologies for which at least one of the following problems occurs: the ontology is impossible to download; the .owl file is corrupted when downloaded; the file is not parseable; the ontology is inconsistent. Furthermore, due to time constraints, we exclude from this preliminary investigation all ontologies whose size exceeds $10,000$ axioms.

This selection results in a corpus of $156$ ontologies, which greatly differ in size and expressivity~\cite{HoPS11}, as summarized in
Table~\ref{tab:summary-ontologies}. For a full list of the corpus, please refer to the
\ifreport
  Appendix.
\else
  technical report \textbf{*** TODO ***}.
\fi

\begin{table}\label{tab:summary-ontologies}
\begin{small}
\begin{center}
\begin{tabular}[ht]{@{~}l@{~~~}c@{~~~}c@{~~~}c@{~}}
\hline\rule{0pt}{4pt}%
Repository  & Range of expressivity &  Range \#axs. & Range sig.\ size   \\[1pt]
\hline\rule{0pt}{4pt}%
BioPortal &\DL{ALCN}-\DL{SHIN(D)/SOIN(D)}	&38--4,735 & 21--3,161\\
TONES		&\DL{AL}-\DL{SROIF(D)/SHOIQ(D)}	&	13--9,629& 14--9,221\\
\hline
\end{tabular}
\end{center}
\end{small}
\caption{Ontology corpus}
\end{table}

\paragraph*{Comparing Syntactic and Semantic Locality.}\label{subsec:design-comparison}

In order to compare syntactic and semantic locality, we want to understand: 
\begin{enumerate}
\item whether, for a given seed signature $\Sigma$, the semantic  $\Sigma$-module is likely to be smaller than the syntactic  $\Sigma$-module, and if so by how much,\footnote{Recall that the semantic  $\Sigma$-module is always a subset of the syntactic  $\Sigma$-module.}
\item how feasible the extraction of semantic modules is.
\end{enumerate}

Here, we focus on the two corresponding notions of $\emptyset$-semantic locality and $\bot$-syntactic locality. In particular, $\bot$-syntactic locality has been throughly investigated in previous work~\cite{DGK+11}, and it has proven to have many interesting properties. A completion of the investigation described in this paper for all fundamental notions of modules is planned in our future work.


Due to the recursive nature of the locality-based module extraction algorithm, we want to investigate locality both on  a
\begin{itemize}
\item per-axiom basis: given an axiom $\alpha$ and a signature $\Sigma$, is it likely that $\alpha$ is semantically $\emptyset$-local w.r.t.\  $\Sigma$ but not syntactically $\bot$- local w.r.t.\  $\Sigma$?
\item per-module basis:  given a signature $\Sigma$, is it likely that $\botmod(\Sigma,\ontol) \neq \sembotmod(\Sigma,\ontol)$? If yes, is it likely that the difference is large? 
\end{itemize}

Hence we need to pick, for each ontology in our corpus, a suitable set of signatures, and this poses a significant problem. First, we do not yet have enough insight into what typical seed signatures are for module extraction. One could assume that large ones are rarely relevant for module extraction---why bother with extracting a large module---but this still leaves a large, i.e., exponential space of possible seed signatures. If $m=\#\Sig{\ontol}$, there are $2^m$ possible seed signatures for which axioms can be tested for locality and for which modules can be extracted. Hence a full investigation is infeasible.

One could assume that the comparison between semantic and syntactic modules could be easier since many signatures can lead to the same module. In other
words, the statistically significant number of modules w.r.t.\ the total number of modules is not larger than that of seed signatures needed w.r.t.\ the total
number of seed signatures. In previous work~\cite{DPSS10womo,DPSS11womo}, however, modules have been studied with respect to how numerous
they are in real-world ontologies. The experiments carried out suggest that the number of modules in ontologies is, in general, exponential w.r.t.\ the size of
the ontology. Moreover, the extraction of enough \emph{different} modules can be hard, because by looking just at seed signatures there is no chance to
avoid the extraction of the same module many times. In particular, for a module $\module$ there can be exponentially many seed signatures w.r.t.\ $\#
\Sig{\module}$ that generate $\module$~\cite{DGK+11}.


As a consequence, we compare the two kinds of locality of axioms---both on a per-axiom basis and a per-module basis---w.r.t.\ random signatures. To avoid any bias, we select a random signature as follows: we set each named entity $\conce$ in the ontology to have probability $p=1/2$ of being included in the signature. Thus each seed signature has the same probability to be chosen. For ontologies whose signature exceeds $9$ entities, in order to get results where 
the true proportion of differences between the two notions of locality lies in the confidence interval $(\pm5\%)$ with confidence level $95\%$, we have to select only $400$ random signatures~\cite{Smi03}. That is, we need to test only  $400$ random signatures  to have a confidence of $95\%$ $(\pm5\%)$ that the differences/equalities we observe reflect the real ones.

\paragraph*{Non-random seed signatures.}

A module, in general, does not necessarily show any internal coherence: intuitively, if we had an ontology describing some knowledge from both the
domains of Geology and of Philosophy, we could still extract the module for the signature $\Sigma=\{\mathtt{Epistemology, Mineral}\}$. This
module is likely to be the union of the two disjoint modules for $\Sigma_1=\{\mathtt{Epistemology}\}$ and $\Sigma_2=\{\mathtt{Mineral}\}$.
This combinatorial behaviour can lead to exponentially many modules in the size of the signature of the ontology and indeed, as mentioned above, the number
of modules in ontologies seems to be exponential~\cite{DPSS10womo,DPSS11womo}.

In contrast to \emph{general} modules, \emph{genuine modules} can be called coherent: they are defined as those modules that cannot be
decomposed into the union of two different modules. Notably, there are only linearly many genuine modules in the size of the ontology $\ontol$, and the set
of genuine modules is a base for all general modules: any module is either genuine or the union of genuine modules.
The linear bound on the number of genuine modules is due to the fact that, for each genuine $x$-module \module, there is an axiom $\alpha$ such that  $\module = \xmod(\tilde\alpha,\ontol)$. 

Thus genuine modules can be said to be interesting modules that we can fully investigate. Hence in addition to the above mentioned investigation of $\bot$- and $\emptyset$-modules for random signatures, we also look at all axiom signatures. 

\medskip

In summary, we test:  
\begin{description}
  \label{desc:tests}
  \item[(T1)]
    for random seed signatures $\Sigma$,
    \begin{description}
      \item[(a)]
        for each axiom $\alpha$ in our corpus, is $\alpha$ semantically $\emptyset$-local w.r.t.\  $\Sigma$ but not syntactically $\bot$- local w.r.t.\  $\Sigma$?
      \item[(b)]
        is $\botmod(\Sigma,\ontol) \neq \sembotmod(\Sigma,\ontol)$? If yes, we determine the difference and its size.
    \end{description}
  \item[(T2)]
    for each axiom signature from our corpus, is  $\botmod(\tilde\alpha,\ontol) \neq \sembotmod(\tilde\alpha,\ontol)$? If yes, we determine the difference and its size.
\end{description}

\comments{
\paragraph{Signatures-to-modules} The description of this challenge in the two cases for modules vs.\ non-local axioms is similar, so we just describe the
first case.

Both sem- and synt-modules are extracted given a seed signature. However, there can be different signatures $\Sigma_1$ and $\Sigma_2$ that lead to the
same module in both cases. Now, we know that synt-modules are an approximation of sem-modules, t.i., given the same signature $\Sigma$, the
sem-module $\module^{\textit{sem}}_\Sigma$ is always contained in the synt-module $\module^{\textit{synt}}_\Sigma$. Then, because of the
observation discussed before, we cannot compare modules only on the base of their axioms, but we need to keep track (in many cases) of the seed
signature of input. In particular, we can have $4$ cases:
\begin{enumerate}
\item $\Sigma_1$ and $\Sigma_2$ lead to the same sem-module $\module^{\textit{sem}}$ as well as to the same synt-module
$\module^{\textit{synt}}$
\item $\Sigma_1$ and $\Sigma_2$ lead to the same sem-module $\module^{\textit{sem}}$ but to $2$ different synt-modules
$\module^{\textit{synt}}_1$ and $\module^{\textit{synt}}_2$
\item $\Sigma_1$ and $\Sigma_2$ lead to $2$ different sem-modules $\module^{\textit{sem}}_1$ and $\module^{\textit{sem}}_2$ but to the same
synt-module $\module^{\textit{synt}}$
\item $\Sigma_1$ and $\Sigma_2$ lead to $2$ different sem-modules $\module^{\textit{sem}}_1$ and $\module^{\textit{sem}}_2$ as well as to $2$
different synt-modules $\module^{\textit{synt}}_1$ and $\module^{\textit{synt}}_2$
\end{enumerate}

For our investigation on the comparison between modules, we create a list of XXXX $\sim 2,000$ triples $(\Sigma, \module^{\textit{sem}}_\Sigma,
\module^{\textit{synt}}_\Sigma)$, where we discard all triples $(\Sigma', \module^{\textit{sem}}_{\Sigma'}, \module^{\textit{synt}}_{\Sigma'})$ if it
occurs that $\module^{\textit{sem}}_\Sigma= \module^{\textit{sem}}_{\Sigma'}$ and $\module^{\textit{synt}}_\Sigma=
\module^{\textit{synt}}_{\Sigma'}$.

\paragraph{Selection of random signatures} 
XXXX For an ontology $\ontol$ whose signature has size $m=\#\Sig{\ontol}$, we know that the most of seed signatures are of size $\sim m/2$. In order
to randomly traverse signatures of all possible sizes, we proceed as follows:
\begin{enumerate}
\item we select a random number $\kappa\in\{1,\ldots, m\}$
\item we generate a random signature $\Sigma\subseteq\Sig{\ontol}$ such that $\#\Sigma=\kappa$.
\end{enumerate}
We repeat this generation until we get XXXX $\sim 2,000$ different triples as described above.
}

\section{Experimental comparison}

\paragraph*{No differences.}
The main result of the experiment is that, for 151 of the 156 ontologies we tested,
no difference between $\bot$- and $\sembot$-locality can be observed.
These 151 ontologies exclude the two NCBO BioPortal ontologies \EFO (Experimental Factor Ontology) and \SWO (Software Ontology),
as well as \Koala, \miniTambis, and \Tambis. 
More specifically,
for every generated seed signature, the corresponding $\bot$- and $\sembot$-module agree,
and every axiom is either $\bot$- and $\sembot$-local, or neither. This statement applies to all randomly
generated seed signatures as well as for \emph{all} axiom signatures -- which are seed signatures for all genuine modules.
We can therefore draw the following conclusions for the 151 ontologies with respect to (T1) and (T2) above.
\begin{description}
  \item[(T1)]
    Given an arbitrary seed signature $\Sigma$,
    there is no difference (a) between $\bot$- and $\sembot$-locality of any given axiom w.r.t.\ $\Sigma$
    and (b) between the $\bot$- and $\sembot$-modules for $\Sigma$, both times at a significance level of $0.05$.
  \item[(T2)]
    Given \emph{any} axiom signature $\Sigma$,
    there is no difference between the $\bot$- and $\sembot$-modules for $\Sigma$.
\end{description}

In the case of the 151 ontologies, the extraction of a $\sembot$-module (with tautology tests performed by FaCT++)
often took considerably longer than the extraction
of the corresponding $\bot$-module. For example, for \Realont{MoleculeRole}, the largest of the 151 ontologies,
times to extract a $\bot$-module
(test all axioms for $\bot$-locality, respectively) ranged between 27 and 169ms (21 and 77ms, respectively),
while the extraction of a $\sembot$-module (test of all axioms for $\sembot$-locality, resp.)
took up to 6\,$\times$ as long, on average 2.7\,$\times$ (2.0\,$\times$, resp.).
It is also worth noting that the ontologies \Galen and \People,
which are renowned for having particularly large $\bot$-modules \cite{CHKS08,DPSS11womo},
are among those without differences between $\bot$- and $\sembot$-locality.

\paragraph*{Differences.}
For the five ontologies where differences between $\bot$- and $\sembot$-modules (or -locality) occur,
we isolated two types of culprits -- axioms which
are not $\bot$-local w.r.t.\ some signature $\Sigma$, but which are $\sembot$-local
w.r.t.\ $\Sigma$.
Type-1 culprits are simple tautologies that have accidentally entered the ``inferred view'' -- i.e., closure under certain entailments --
of two ontologies. They do not occur in the original ``asserted'' versions and can, in principle, be detected by a slightly refined
syntactic locality check.
Type-2 culprits are definitions of concept names via a conjunction
that satisfies certain conditions explained below.
There are not many type-1 and type-2 axioms in the affected ontologies, and the observed differences are comparably small.
Table \ref{tab:overview_differences} gives an overview of the differences observed.
\begin{table}
  \centering
  \begin{tabular}{@{~~~}l@{~~}r@{~~~}l@{~~}r@{~~~}r@{~~}r@{~~~}r@{~~~}l@{~~~}}
    \hline\rule{0pt}{11pt}%
    Ontology    & \#axs    & \multicolumn{2}{l@{~~}}{\#differences} & \multicolumn{2}{l@{~~}}{difference} & time   & culprit       \\
                &          &         &                              & \multicolumn{2}{l@{~~}}{sizes}      & ratio  & type and      \\
                &          &         &                              & \#axs   & rel.                      & avg.\  & frequency     \\[1pt]
    \hline\rule{0pt}{11pt}%
    \SWO        &     3446 & T1\,a   & 400                          & 6--22   & 0--1\%                    & 3.31   & 1 (30$\times$)  \\
                &          & T1\,b   & 400                          & 23--29  & 1--2\%                    & 5.11   &                 \\
                &          & T2      & 3446                         & 3--1    & 1--5\%                    & 5.86   &                 \\[4pt]
    \EFO        &     6008 & T1\,a   & 400                          & 8--24   & 0--1\%                    & 1.42   & 1 (32$\times$)  \\
                &          & T1\,b   & 400                          & 13--30  & 0--1\%                    & 1.38   &                 \\
                &          & T2      & 128                          & 1--4    & 9--17\%                   & ---    &                 \\[4pt]
    \Koala      &       42 & T1\,a   &   0                          &     0   & 0\%                       & ---    & 2  (1$\times$)  \\
                &          & T1\,b   &   2                          &     1   & 3\%                       & ---    &                 \\
                &          & T2      &   0                          &     0   & 0\%                       & ---    &                 \\[4pt]
    \miniTambis &      170 & T1\,a   &  68                          &  1--2   & 1--3\%                    & ---    & 2  (3$\times$)  \\
                &          & T1\,b   &  93                          &  1--4   & 1--3\%                    & ---    &                 \\
                &          & T2      &  26                          &  1--7   & 6--75\%                   & ---    &                 \\[4pt]
    \Tambis     &      592 & T1\,a   &  58                          &  1--3   & 0--1\%                    & 3.31   & 2 (11$\times$)  \\
                &          & T1\,b   &  229                         &  2--11  & 0--2\%                    & 5.01   &                 \\
                &          & T2      &  191                         &  4--41  & 2--26\%                   & ---    &                 \\[1pt]
    \hline\rule{0pt}{11pt}%
  \end{tabular}
  \caption{%
    Overview table of differences observed. The columns show: the ontology name; the overall number of axioms;
    the name of the test (see list on Page \pageref{desc:tests}); the number of cases with differences;
    the number of axioms in the differences (absolute and relative to the $\bot$-case);
    the average time ratio $\sembot:\bot$ (``---'' indicates that no reliable statement is possible:
    the time for $\bot$ is only a few, often 0, milliseconds);
    the type of culprit present and the number of axioms of this type.
  }
  \label{tab:overview_differences}
\end{table}

\paragraph*{Type-1 culprits} are axioms 
  \texttt{InverseObjectProperties(P, InverseOf(P))},\linebreak
where \texttt{P} is a role. This translates into the tautology $\texttt{P} \equiv (\texttt{P}^-)^-$ in DL notation.
Such an axiom is therefore $\sembot$-local w.r.t.\ any signature.
However, it behaves differently for $\bot$-locality: if the signature $\Sigma$ contains \texttt{P},
then both sides of the equation are neither in $\Bot(\Sigma)$ nor in $\Top(\Sigma)$, hence the axiom is considered non-local;
otherwise, both sides are $\bot$-equivalent, hence the axiom is local.

Type-1 axioms occur in the ``inferred view'' of the ontologies \EFO and \SWO.
Table \ref{tab:overview_differences} shows the relatively modest differences caused by these axioms.
In all cases, there are no other axioms in the differences.
This means that no differences occur for the non-inferred original versions of \EFO and \SWO.

\paragraph*{Type-2 culprits} are complex definitions $A \equiv C$ of a concept name $A$
where $C$ is a disjunction that contains both a universal and an existential (or minimum cardinality) restriction on the same role.
This affects the ontologies \Koala, \miniTambis, and \Tambis. 
The effect is best illustrated for \Koala,
which contains exactly one such axiom, namely
$
  \texttt{M} ~~\equiv~~
    \texttt{S} ~\sqcap~ \forall\texttt{c}.\texttt{F}
    ~\sqcap~ \forall\texttt{g}.\{\texttt{m}\}
    ~\sqcap~ \mathord{=}3\,\texttt{c}.\top,
$
where we have abbreviated the concept names \texttt{\underline{M}aleStudentWith3Daughters}, \texttt{\underline{S}tudent}, \texttt{\underline{F}emale},
the roles \texttt{has\underline{C}hildren}, \texttt{has\underline{G}ender}, and the nominal \texttt{\underline{m}ale}.
Now if the signature against which the axiom is tested for locality
contains $\{\texttt{S},\texttt{c},\texttt{g}\}$
but neither \texttt{M} nor \texttt{F},
then this axiom is not $\bot$-local because none of the conjuncts on the right-hand side is in $\Bot(\Sigma)$.
On the other hand, this axiom is a tautology when \texttt{M} and \texttt{F}
are replaced by $\bot$: the conjunction $\forall\texttt{c}.\bot \sqcap \mathord{=}3\,\texttt{c}.\top$
cannot have any instances, regardless of how $\texttt{c}$ is interpreted.

For \Koala, this effect only causes two singleton differences between sets of local axioms for the randomly generated seed signatures,
as shown in Table \ref{tab:overview_differences}. For axiom signatures,
there is no difference. Interestingly, this effect does not propagate to modules: for all signatures,
$\bot$- and $\sembot$-modules are the same. The reason might be that (a) \texttt{g} is used in many axioms
and is thus very likely to contribute to the extended signature during module extraction,
and (b) then the axiom defining \texttt{F} is no longer local, which ``pulls'' \texttt{F} into the extended signature,
preventing the observed effect.

In \miniTambis and \Tambis, this effect is much stronger and affects a large proportion of modules, as shown in~Table \ref{tab:overview_differences}.
The differences in these cases do not only consist of culprit axioms, but also of axioms that become non-local
after the signature has been extended by the terms in the culprit axioms. Still, the size of the differences
is mostly modest while, for \Tambis, the $\sembot$-locality test ($\sembot$-module extraction) takes
on average over three times (five times) as long as the $\bot$-locality test ($\bot$-module extraction).

\section{Conclusion and Outlook}\label{sec:conclusion}

\paragraph*{Summary.}

We obtain two main observations from the experiments carried out.
\begin{itemize}
  \item
    In practice, there is no or little difference between semantic and syntactic locality.
    That is, the computationally cheaper
    syntactic locality is a good approximation of
    semantic locality.
  \item
    Though in principle hard to compute, semantic modules can be extracted rather fast in practice.
\end{itemize}

These results suggest that it is questionable to conclude that semantic locality should be preferred to syntactic locality.
In terms of computation time,
there is often a benefit in using syntactic locality: the average speed-up compared to the extraction of a semantic-locality based module
is by a factor of up to 6. For some particular module pairs, it is higher by an order of magnitude. The gain in module size is zero or so small
that it is hard to justify the extra time spent.
In particular, there is no gain in size for the
ontologies \Galen and \People, which are ``renowned'' for having disproportionately large modules \cite{CHKS08,DPSS11womo}.

Our results are interesting not only because they provide an evaluation of how good the cheap syntactic locality approximates semantic
locality, but also because they enabled us to fix bugs in the implementation of syntactic modularity. For example, earlier data from the experiment
have shown that reflexivity axioms had been treated incorrectly by the syntactic locality checker.

\paragraph*{Future Work.}
It is evident that this work is preliminary.
It investigates only the differences between the related notions of
$\bot$- and $\sembot$-locality.
We plan to extend the same study to other notions of locality, in particular,
nested modules ($\tbstar$- vs.\ $\semtop\sembot^*$-modules) -- these notions are the most economical in terms of module size.
Moreover, we want to extend the investigation to the remaining larger ontologies in the BioPortal repository
and further large ontologies, e.g., some versions of the NCI Thesaurus\footnote{Downloadable from
\url{http://evs.nci.nih.gov/ftp1/NCI_Thesaurus}}.
Preliminary results with a version that is not among the regular releases
show differences due to type-2 culprits, but we have not included them here
because the differences disappear after removing axioms that were introduced due
a problem with object and annotation properties when the ontology file is parsed by the OWL API.
This behaviour is yet to be investigated and explained.

Another interesting extension is to modify the seed signature sampling.
Currently, the random variable ``size of the seed signature generated'' follows the binomial distribution
with expected value $m/2$ and variance $m/4$.
Hence, most signatures in the sample have size around $m/2$;
small and large signatures are underrepresented.
For example, for one ontology with $915$ terms,
all signature sizes lay between $422$ and $509$.
One might argue that, for big ontologies, the typical module extraction scenario
does not require large seed signatures -- but it does sometimes require relatively small seed signatures,
for example, when a module is extracted to efficiently answer a given entailment query of
typically small size.
On the other hand, large modules resulting from larger seed signatures may be more likely
to differ.
We therefore plan an alternative seed signature sampling via bins for average signature sizes:
repeat the current sampling procedure scaled to several subintervals of the range of possible signature sizes.

\comments{However, the typical random signature has size close to $\sim m/2$. In order to possibly traverse signatures of all sizes, then,
for each ontology $\ontol$ we extract $m$ modules, whose each seed signature is generated as described in what follows:
\begin{itemize}
\item[-] we generate a random number $\kappa\in\{1,\ldots,m\}$
\item[-] we extract the first $\kappa$ elements from a random list of the ontology's signature.
\end{itemize}
}

Our current results answer the question whether there is a significant difference
between the two locality notions \emph{with respect to a given signature}.
It is also interesting to ask the same question relative to a given module.
To answer it, the sampling of modules instead of seed signatures requires further investigation.

\paragraph*{Acknowledgment.} We thank Rafael Gon\c{c}alves \ifreport\else and the anonymous reviewers \fi for helpful comments.

\bibliographystyle{splncs03}
\bibliography{short-string,biblio-clean}

\ifreport
  \newpage
  \appendix

  \section*{Appendix: overview of the ontologies used}

  \begin{small}
  \begin{center}
  \begin{longtable}[ht]{llr@{~~}r}
  \hline\rule{0pt}{9pt}%
  Ontology  & Expressivity &  \#Axioms & Sig.\ size   \\[1pt]
  \hline
  \endfirsthead
  \multicolumn{4}{c}%
  {\tablename\ \thetable\ -- \textit{Continued from previous page}} \\
  \hline\rule{0pt}{9pt}%
  Ontology  & Expressivity &  \#Axioms & Sig.\ size   \\[1pt]
  \hline
  \endhead
  \hline \multicolumn{4}{r}{\textit{Continued on next page}} \\
  \endfoot
  \hline
  \endlastfoot

  \Realont{aba-adult-mouse-brain}&\DL{ALCI}&3,441&915\\
  \Realont{adverse-event-reporting-ontology}&\DL{SHOIN(D)}&574&503\\
  \Realont{african-traditional-medicine}&\DL{ALE}&208&225\\
  \Realont{amino-acid}&\DL{ALCF(D)}&477&52\\
  \Realont{amphibian-gross-anatomy}&\DL{ALE}&2,673&1,647\\
  \Realont{anatomical-entity-ontology}&\DL{ALE}&352&359\\
  \Realont{ascomycete-phenotype-ontology}&\DL{AL}&294&329\\
  \Realont{basic-formal-ontology}&\DL{ALC}&95&39\\
  \Realont{basic-vertebrate-anatomy}&\DL{SHIF}&388&231\\
  \Realont{bilateria-anatomy}&\DL{ALEH+}&138&121\\
  \Realont{bioinformatics-data-formats-identifiers...}&\DL{ALE+}&3,803&2,844\\
  \Realont{biological-imaging-methods}&\DL{S}&548&626\\
  \Realont{biomedical-resource-ontology}&\DL{SHIF(D)}&681&672\\
  \Realont{biopax}&\DL{SHIN(D)}&391&165\\
  \Realont{biotop}&\DL{SRI}&680&404\\
  \Realont{birnlex}&\DL{AL}&3,572&3,589\\
  \Realont{bleeding-history-phenotype}&\DL{ALCIF(D)}&1,925&582\\
  \Realont{body-system}&\DL{AL}&28&30\\
  \Realont{breast-tissue-cell-lines}&\DL{ALCH(D)}&2,734&412\\
  \Realont{brenda-tissue-enzyme-source}&\DL{ALE}&6,284&5,272\\
  \Realont{c-elegans-development}&\DL{AL}&71&73\\
  \Realont{c-elegans-phenotype}&\DL{AL+}&2,279&2,026\\
  \Realont{cao}&\DL{SHIQ(D)}&476&290\\
  \Realont{cell-behavior-ontology}&\DL{ALUO}&13&14\\
  \Realont{cell-type}&\DL{ALC}&2,975&2,012\\
  \Realont{cereal-plant-development}&\DL{ALE}&235&237\\
  \Realont{cereal-plant-gross-anatomy}&\DL{ALE+}&1,839&1,173\\
  \Realont{cognitive-atlas}&\DL{ALC}&3,622&1,585\\
  \Realont{common-anatomy-reference-ontology}&\DL{ALE+}&54&54\\
  \Realont{common-terminology-criteria-for-adverse...}&\DL{AL(D)}&6,940&3,889\\
  \Realont{dendritic-cell	}&\DL{ALC}&313&192\\
  \Realont{dikb-evidence-ontology}&\DL{ALCHOIN(D)}&640&251\\
  \Realont{drosophila-development}&\DL{ALEH+}&410&138\\
  \Realont{electrocardiography-ontology}&\DL{ALCIF(D)}&1,274&1,171\\
  \Realont{environment-ontology}&\DL{S}&1,807&1,574\\
  \Realont{epilepsy}&\DL{ALH(D)}&145&148\\
  \Realont{event-inoh-pathway-ontology}&\DL{ALEH+}&7,131&3,836\\
  \Realont{evidence-codes}&\DL{ALE}&342&268\\
  \Realont{exo}&\DL{ALE+}&85&121\\
  \Realont{experimental-factor-ontology}&\DL{ALHIF+}&6,008&4,869\\
  \Realont{fda-medical-devices-2010}&\DL{AL}&4,907&4,941\\
  \Realont{fly-taxonomy}&\DL{AL}&6,587&6,599\\
  \Realont{flybase-controlled-vocabulary}&\DL{ALE+}&659&771\\
  \Realont{fungal-gross-anatomy}&\DL{ALEI+}&106&86\\
  \Realont{gene-regulation-ontology}&\DL{ALCHIQ(D)}&962&544\\
  \Realont{general-formal-ontology}&\DL{SHIQ}&212&86\\
  \Realont{hom-datasource\_oshpd}&\DL{AL}&351&361\\
  \Realont{hom-datasource\_oshpdsc}&\DL{AL}&351&360\\
  \Realont{hom-dxprocs\_mdcdrg}&\DL{AL}&774&784\\
  \Realont{hom-harvard}&\DL{AL}&189&191\\
  \Realont{hom-icd9\_procs\_oshpd}&\DL{AL}&4,642&4,652\\
  \Realont{hom-icd9cm-ecodes}&\DL{AL}&1,490&1,500\\
  \Realont{hom-icd9cm\_procedures}&\DL{AL}&4,644&4,656\\
  \Realont{hom-mdcdrg\_oshpd}&\DL{AL}&773&782\\
  \Realont{hom-oshpd-sc}&\DL{AL}&266&278\\
  \Realont{hom-oshpd\_usecase}&\DL{AL}&393&408\\
  \Realont{hom-procs2\_oshpd}&\DL{AL}&4,642&4,652\\
  \Realont{hom-ucare}&\DL{AL}&64&75\\
  \Realont{hom\_mdcs-drgs}&\DL{AL}&774&780\\
  \Realont{homerun-ontology}&\DL{AL}&1,194&1,094\\
  \Realont{host-pathogen-interactions-ontology}&\DL{SHI}&403&319\\
  \Realont{human-developmental-anatomy-abs...}&\DL{ALE}&2,335&2,316\\
  \Realont{human-developmental-anatomy-tim...}&\DL{ALE}&8,339&8,343\\
  \Realont{human-disease}&\DL{AL}&6,753&8,625\\
  \Realont{hymenoptera-anatomy-ontology}&\DL{SR}&8,493&4,324\\
  \Realont{imgt-ontology}&\DL{ALCIN(D)}&1,112&122\\
  \Realont{infectious-disease-ontology}&\DL{SROIF}&1,221&640\\
  \Realont{information-artifact-ontology}&\DL{SHOIN(D)}&294&197\\
  \Realont{interaction-network-ontology}&\DL{ALC}&1,034&981\\
  \Realont{ixno}&\DL{AL}&39&53\\
  \Realont{leukocyte-surface-markers}&\DL{AL+}&472&473\\
  \Realont{linkingkin2pep}&\DL{SHIF(D)}&30&17\\
  \Realont{lipid-ontology}&\DL{ALCHIN}&2,375&762\\
  \Realont{loggerhead-nesting}&\DL{ALE}&347&314\\
  \Realont{maize-gross-anatomy}&\DL{ALE}&217&184\\
  \Realont{mass-spectrometry}&\DL{SH}&4,447&4,492\\
  \Realont{medaka-fish-anatomy-and-dev...}&\DL{ALE}&4,402&4,363\\
  \Realont{mego}&\DL{ALE+}&421&370\\
  \Realont{minimal-anatomical-terminology}&\DL{ALE}&504&481\\
  \Realont{molecule-role-inoh-protein-name...}&\DL{ALE+}&9,629&9221\\
  \Realont{mouse-adult-gross-anatomy}&\DL{ALE+}&3,776&2,984\\
  \Realont{mouse-pathology}&\DL{ALE+}&808&757\\
  \Realont{multiple-alignment}&\DL{ALE+}&168&174\\
  \Realont{neomark-oral-cancer-ontology}&\DL{SHIQ}&399&352\\
  \Realont{neural-electromagnetic-ontologies}&\DL{SHIQ(D)}&2,578&1766\\
  \Realont{neural-immune-gene-ontology}&\DL{SH}&8,835&4,843\\
  \Realont{neuro-behavior-ontology}&\DL{AL}&768&733\\
  \Realont{nif-dysfunction}&\DL{SROIF(D)}&2,635&2,951\\
  \Realont{nmr-instrument-specific-component...}&\DL{AL}&290&301\\
  \Realont{obo-relationship-types}&\DL{ALR+}&33&26\\
  \Realont{ontology-for-drug-discovery-investigations}&\DL{SHOIN(D)}&996&837\\
  \Realont{ontology-for-general-medical-science}&\DL{ALCO}&216&162\\
  \Realont{ontology-for-genetic-interval}&\DL{SHIN(D)}&509&298\\
  \Realont{ontology-for-microrna-target-prediction}&\DL{ALCI(D)}&415&338\\
  \Realont{ontology-for-parasite-lifecycle}&\DL{SHOIF}&855&415\\
  \Realont{ontology-of-general-purpose-datatypes}&\DL{ALCHOI}&459&193\\
  \Realont{ontology-of-geographical-region}&\DL{AL}&38&39\\
  \Realont{ontology-of-glucose-metabolism-disorder}&\DL{AL}&132&132\\
  \Realont{ontology-of-medically-related-social-entities}&\DL{ALCO}&157&99\\
  \Realont{ontology-of-physics-for-biology}&\DL{ALCHIQ(D)}&795&545\\
  \Realont{pathogen-transmission}&\DL{AL}&24&28\\
  \Realont{pediatric-terminology}&\DL{AL}&894&891\\
  \Realont{phare}&\DL{ALCHIF(D)}&459&312\\
  \Realont{phenotypic-quality}&\DL{SH}&1,831&2,282\\
  \Realont{phylogenetic-ontology}&\DL{AL}&77&83\\
  \Realont{physicalfields}&\DL{ALI}&136&78\\
  \Realont{physico-chemical-process}&\DL{ALE}&734&560\\
  \Realont{pilot-ontology}&\DL{ALCIF(D)}&85&39\\
  \Realont{pko\_re}&\DL{ALCF}&771&770\\
  \Realont{plant-environmental-conditions}&\DL{AL}&499&501\\
  \Realont{plant-growth-and-development-stage}&\DL{ALE+}&240&285\\
  \Realont{plant-ontology}&\DL{S}&2,215&1,460\\
  \Realont{plant-trait-ontology}&\DL{ALE}&1,290&1,124\\
  \Realont{platynereis-stage-ontology}&\DL{ALE}&31&18\\
  \Realont{protein-modification}&\DL{ALE+}&1,986&1,346\\
  \Realont{protein-ontology}&\DL{ALCF(D)}&689&226\\
  \Realont{protein-protein-interaction}&\DL{ALE+}&1,007&962\\
  \Realont{pseudogene}&\DL{AL}&19&23\\
  \Realont{quantitative-imaging-biomarker...}&\DL{ALUIF(D)}&1,697&1,381\\
  \Realont{rat-strain-ontology}&\DL{ALE}&4,122&3,004\\
  \Realont{reproductive-trait-and-phenotype...}&\DL{AL}&91&96\\
  \Realont{sample-processing-and-sep...}&\DL{AL}&193&194\\
  \Realont{sequence-types-and-features}&\DL{SHI}&2,545&2,167\\
  \Realont{sleep-domain-ontology}&\DL{SHIF(D)}&363&256\\
  \Realont{smoking-behavior-risk-ontology}&\DL{ALEI+}&185&135\\
  \Realont{software-ontology}&\DL{SHOIQ(D)}&3,446&1,039\\
  \Realont{spatial-ontology}&\DL{ALEHI+}&235&172\\
  \Realont{spider-ontology}&\DL{ALE+}&778&581\\
  \Realont{student-health-record}&\DL{ALH(D)}&418&382\\
  \Realont{symptom-ontology}&\DL{AL}&839&935\\
  \Realont{syndromic-surveillance-ontology}&\DL{ALIF(D)}&1,679&364\\
  \Realont{sysmo-jerm}&\DL{SI(D)}&417&280\\
  \Realont{systems-biology}&\DL{AL}&587&558\\
  \Realont{systems-chemical-biology-chemogenomics}&\DL{SHIN(D)}&489&216\\
  \Realont{taxonomic-rank-vocabulary}&\DL{AL}&58&59\\
  \Realont{tick-gross-anatomy}&\DL{ALE+}&948&630\\
  \Realont{tissue-microarray-ontology}&\DL{ALI(D)}&60&32\\
  \Realont{tok\_ontology}&\DL{SRIQ(D)}&466&331\\
  \Realont{translational-medicine-ontology}&\DL{SRIN(D)}&499&389\\
  \Realont{units-of-measurement}&\DL{ALE}&343&336\\
  \Realont{units-ontology}&\DL{SHIF}&105&88\\
  \Realont{vertebrate-anatomy-ontology}&\DL{ALER+}&340&234\\
  \Realont{vertebrate-homologous-organ-groups}&\DL{ALE+}&1,689&1,186\\
  \Realont{vertebrate-trait-ontology}&\DL{AL+}&3,586&3,072\\
  \Realont{web-service-interaction-ontology}&\DL{ALER+}&29&39\\
  \Realont{wheat-trait}&\DL{AL}&175&176\\
  \Realont{xenopus-anatomy-and-development}&\DL{ALE+}&2,243&1,051\\
  \Realont{yeast-phenotypes}&\DL{AL}&266&300\\
  \hline
  \end{longtable}
  \end{center}
  \end{small}
\fi

\end{document}